\documentclass[journal, 10pt, twocolumn, twoside]{IEEEtran}
\usepackage{color}
\usepackage{multirow}
\usepackage{cite}

  \usepackage[pdftex]{graphicx}
\usepackage[cmex10]{amsmath}
\usepackage{amsfonts}
\usepackage{amssymb}
\usepackage{algorithmic}
  \usepackage[caption=false,font=scriptsize]{subfig}

\begin{document}
%
\title{\huge Dictionary Integration using 3D Morphable Face Models for Pose-invariant Collaborative-representation-based Classification}
%
%
%

\author{Xiaoning Song,
	Zhen-Hua~Feng~\IEEEmembership{Member,~IEEE},
	Guosheng Hu,
    Josef~Kittler~\IEEEmembership{Life~Member,~IEEE},
	William~Christmas,
	and~Xiao-Jun~Wu
\thanks{This work was partially supported by the Engineering and Physical Sciences Research Council Programme Grants `FACER2VM' (EP/N007743/1), the National Natural Science Foundation of China (Grant No. 61672265), the Natural Science Foundation of Jiangsu Province (Grant No. BK20161135), China Postdoctoral Science Foundation (Grant No. 2016M590407), the Fundamental Research Funds for the Central Universities (Grant No. JUSRP115A29), the Open Project Program of Key Laboratory of Intelligent Perception and Systems for High-Dimensional Information of Ministry of Education (No. JYB201603), the National Science and Technology Support Program of China (Grant No. 2015BAD17B02) and the Priority Academic Program Development (PAPD) of Jiangsu Higer Education Institutions and Jiangsu Collaborative Innovation Center on Atmospheric Environment and Equipment Technology (CICAEET).}
\thanks{X. Song and X.-J. Wu are with the School of Internet of Things, Jiangnan University, Wuxi 214122, China (e-mail: xnsong@hotmail.com, xiaojun\_wu\_jnu@163.com).}
\thanks{Z.-H. Feng, J. Kittler and W. Christmas are with the Centre for Vision, Speech and Signal Processing,
University of Surrey, Guildford GU2 7XH, UK (e-mail: z.feng, j.kittler, w.christmas@surrey.ac.uk).}
\thanks{G. Hu is with Anyvision, Queen’s Road, Belfast BT39DT, UK (e-mail: guosheng.hu@anyvision.co).}
}

\maketitle

\begin{abstract}
The paper presents a dictionary integration algorithm using 3D morphable face models (3DMM) for pose-invariant collaborative-representation-based face classification.
To this end, we first fit a 3DMM to the 2D face images of a dictionary to reconstruct the 3D shape and texture of each image.
The 3D faces are used to render a number of virtual 2D face images with arbitrary pose variations to augment the training data,
by merging the original and rendered virtual samples {\color{black}to create} an extended dictionary. 
Second, to reduce the information redundancy of the extended dictionary and improve the sparsity of reconstruction coefficient vectors using collaborative-representation-based classification (CRC), we exploit an on-line elimination scheme to optimise the extended dictionary by identifying the most representative training samples for a given query. 
The final goal is to perform pose-invariant face classification using the proposed dictionary integration method and the on-line pruning strategy under the CRC framework. 
Experimental results obtained for a set of well-known face datasets demonstrate the merits of the proposed method, especially its robustness to pose variations.
\end{abstract}


%

\section{Introduction}
\label{section_1}
Sparse-representation-based classification (SRC) and collaborative-representation-based classification (CRC) approaches
have introduced a new concept in pattern recognition~\cite{wright2009robust,ortiz2013face,chen2013blessing,yang2010metaface,li2010local,zhang2012collaborative,zhang2011sparse,song2015progressive,song2016towards}.
The aim of SRC or CRC is to represent a new observation, 
also known as a signal or a sample, using a minimal number of training samples selected from an existing dictionary that consists of a number of observations across different classes.
To achieve this objective, the $\ell_1$-norm constraint is used as a regularization term in SRC to obtain sparse reconstruction coefficient vectors.
In contrast, CRC obtains reconstruction coefficient vectors using $\ell_2$-norm regularisation.
It has been proven that the $\ell_2$-norm based regularization of the coefficient vector in CRC helps to achieve competitive accuracy at much lower computational cost than that of the $\ell_1$-norm constraint in SRC~\cite{zhang2012collaborative}.

In SRC and CRC, given a new observation, the task of classification is performed by comparing the capacity of the samples from the individual classes in a training set to represent the new observation.
The decision is made by selecting the label of the class yielding the minimum reconstruction error for the new observation.
The robustness of classification is based on the assumption that we have an over-complete dictionary, \textit{i.e.} an arbitrary observation can be approximated well by a linear combination of finite samples in the dictionary.
However, in some practical scenarios such as CCTV security systems, only a few or even just a single image of a subject is available for training.
Such a dictionary cannot fully reflect the appearance of a query sample, especially in the presence of illumination, expression, occlusion and pose variations.
To address this issue, we explore the use of a 3D morphable face model (3DMM)~\cite{Tena_3DMM,ME_ICB2013,vetter1998synthesis} in generating virtual training samples for pose-invariant CRC-based face classification.

To generate virtual training samples, a widely used method is to perturb original samples to extend the current dataset.
For example, Deng et al. proposed the extended sparse-representation-based classification (ESRC) algorithm that imports an intraclass variation dictionary for under-sampled face recognition~\cite{deng2012extended}. 
Ryu et al. exploited the distribution of the samples in a given gallery set to generate virtual training samples for face recognition, by fusing multiple training samples in the PCA-based feature space~\cite{ryu2002simple}.
Beymer et al. constructed new face images with different poses using an exemplar-based method and improved the accuracy of face recognition~\cite{beymer1995face}.
In facial landmark detection, 
random perturbations are usually applied to initial landmarks to augment the volume of a training dataset for successful landmark detector training~\cite{cootes2001active,cao2014face, feng2015random}.
As another concept, symmetrical faces have been used for data augmentation in face detection and classification in~\cite{su1999application,saha2007symmetry,saber1998frontal}.
Xu et al. proposed to use symmetrical faces in face recognition with a sparse-representation-based method~\cite{xu2013using,xuintegrating}.
\begin{figure}[t]
	\begin{center}
		\includegraphics[trim =5mm 5mm 5mm 5mm, clip, width=1\linewidth]{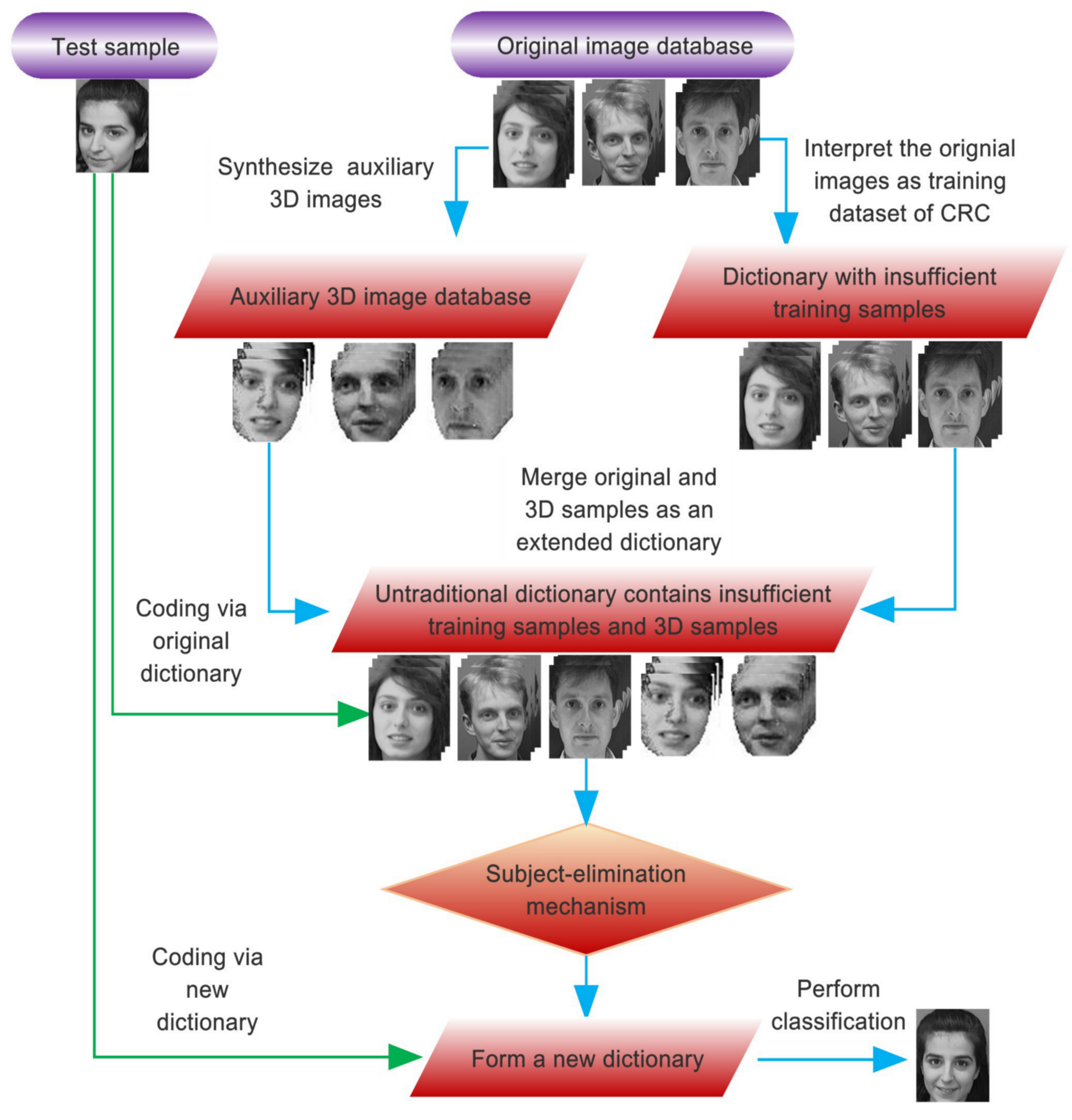}
	\end{center}
	\caption{The schematic diagram of the proposed framework}
	\label{fig1}
\end{figure}

Although the methods mentioned above lead to higher accuracy in face recognition or better performance in other computer vision and pattern recognition tasks, the generated virtual samples cannot tackle the problem of pose variations very well.
The major drawback of traditional virtual sample generation methods is the inability to represent intra-class pose variations adequately. 
To be more specific, if the intra-class pose variations of test samples different from those of the subjects in the gallery set, the information conveyed by the original training samples may not be sufficient to reconstruct {\color{black} them}.
Even an extended dictionary consisting of both original and virtual samples often lacks the capacity to represent a test face image of arbitrary pose. 
The traditional virtual sample generation methods used to construct an auxiliary dictionary for the relevant types of variations typically ignore the pose differences between gallery and query sets. 
Lastly, a large number of generated virtual samples may lead to information redundancy of the extended dictionary and data uncertainty in decision making. 

To address the above issues, in this paper, we develop a method to extend an existing dictionary using a generative 3DMM.
As compared to 2D generative models such as active appearance models (AAM)~\cite{cootes2001active,feng2012automatic}, a 3DMM is capable of generating diverse face instances with arbitrary pose and illumination variations.
It has been already widely used in some computer vision applications.
For example, Feng et al. used a 3DMM to generate a set of virtual faces for a facial landmark detector training and obtained state-of-the-art detection results for faces in the wild, using a cascaded collaborative regression method~\cite{feng2015cascaded,kittler20163d}.
R\"{a}tsch et al. generated virtual faces using 3DMM for 2D pose estimation using support vector regression~\cite{ratsch_wavelet_2012}.
In this paper, we propose to apply 3DMM to the training images of a given dictionary and synthesise a number of new faces with different pose variations as an auxiliary dictionary.
The extended dictionary obtained using 3DMM generated entries is much better in representing different modes of variations than the original training faces alone.
{\color{black} Moreover, a hypothesis} elimination scheme {\color{black}with the associated on-line dictionary pruning} is jointly used with the CRC method to perform face classification.
Fig.~\ref{fig1} shows the schematic diagram of the proposed framework.
The contributions of our work are three-fold. 
\begin{itemize}
	\item{To obtain an extended dictionary, for each 2D training example, we use a 3DMM fitting algorithm to reconstruct the 3D shape and texture information and render additional face images with pose and potentially illumination variations. The original and rendered virtual faces are used to form the extended dictionary.} 
	\item{To optimise the extended dictionary and address the problem of information redundancy {\color{black} during testing}, we exploit {\color{black} an on-line hypothesis} elimination scheme to discard training samples with inferior representation capabilities.}
	\item{We propose a CRC-based method to perform pose-invariant face classification, by mining the most representative training samples from the dictionary {\color{black} extended} using 3DMM generated faces.
	In the rest of this paper, we use the term `3D Pose Dictionary integration in CRC' (3DPD-CRC) for the proposed algorithm.}
\end{itemize}

The rest of this paper is organised as follows: Section 2 overviews the relevant classical classification algorithms including SRC and CRC.
They are the prerequisites to our method proposed in Section 3. Section 4 presents a theoretical analysis to the proposed method and Section 5 reports the results of comprehensive experiments conducted on the well-known ORL, FERET and PIE face datasets. Lastly, we summarize the paper in Section 6.

\section{Background}
Given a dictionary with $K \times M$ training samples $\{\mathbf{x}_{1,1},...,\mathbf{x}_{K,M}\}$, 
where $K$ is the number of classes and $M$ is the number of training samples from each class, 
a test sample $\mathbf{y} \in \mathbb{R}^P$ can be approximated by the linear combination of all these training samples:
\begin{equation}
	\label{equ1}
	\mathbf{y} \approx \sum_{k=1}^{K}\sum_{m=1}^{M}{\alpha}_{k,m}\mathbf{x}_{k,m},
\end{equation}
where $ {\alpha}_{k,m} $ is the entry of the coefficient vector corresponding to the $m$th training sample in the $k$th class $\mathbf{x}_{k,m} \in \mathbb{R}^P$,
$P$ is the dimensionality of a sample. 
The entry $ {\alpha}_{k,m} $ indicates the potential of the corresponding training sample to represent the test sample $\mathbf{y}$. 
It should be noted that the number of training samples of each class can be varied. Here we just use the same number, $M$, for convenience. 
In addition, Eq.~(\ref{equ1}) can compactly be rewritten as:
\begin{equation}
	\label{equ2}
	\mathbf{y} \approx \mathbf{X}\boldsymbol{\alpha},
\end{equation}
where $ \mathbf{X}=[\mathbf{x}_{1,1},...,\mathbf{x}_{K,M}] \in \mathbb{R}^{P \times KM}$ is the dictionary matrix containing all the training samples 
and $\boldsymbol{\alpha}=[{\alpha}_{1,1},...,{\alpha}_{KM}]^T$  is the coefficient vector need to be estimated.

Once the coefficient vector is obtained, we can measure the propensity of the $k$th class to represent the test sample:
\begin{equation}
	\label{equ5}
	\mathbf{c}_{k} = \sum_{m=1}^{M} \alpha_{k,m}\mathbf{x}_{k,m},
\end{equation}
where $\mathbf{c}_{k}$ is the reconstruction of the test sample using the training samples merely from the $k$th class.
The test sample reconstruction error for the $k$th class is obtained by:
\begin{equation}
\label{equ_reconstuction_error}
	E(\mathbf{y})_k= \parallel \mathbf{y} - \mathbf{c}_k \parallel_2,
\end{equation}
and the label of the test sample $\mathbf{y}$ is determined using:
\begin{equation}
\label{equ11_2}
	Label(\mathbf{y})= \underset{k} {\mathrm{argmin}}  \{E(\mathbf{y})_k\}.
\end{equation}

As stated above, the key to the classification problem is to obtain the coefficient vector reconstructing the test sample.
To solve this problem, in the rest of this section, we briefly overview two algorithms: the sparse-representation-based classification (SRC)~\cite{wright2009robust} and collaborative-representation-based classification (CRC)~\cite{zhang2012collaborative}.

\subsubsection{SRC} 
The aim of SRC is to obtain a sparse coefficient vector $\boldsymbol{\alpha}$ by minimising the objective function:
\begin{align}
	\label{equ0}
	&\min{\parallel\boldsymbol{\alpha} \parallel}_0 \\
	&\text{s.t.    } \hspace{0.2cm} \mathbf{y}  =  \mathbf{X}\boldsymbol{\alpha}. \nonumber
\end{align}
However, this $\ell_0$-norm constrained optimisation problem is NP-hard and difficult to solve.
To address this issue, some recent studies~\cite{wright2009robust,donoho2006most,candes2006stable,candes2006near} demonstrate that if $\boldsymbol{\alpha}$ is sparse enough, the solution to the above problem is equal to the solution of:
\begin{align}
	\label{equ0_1}
	&\min{\parallel\boldsymbol{\alpha} \parallel}_1 \\
	&\text{s.t.    } \hspace{0.2cm} \mathbf{y}  =  \mathbf{X}\boldsymbol{\alpha}. \nonumber
\end{align}
This optimisation problem can be solved by standard linear programming methods in polynomial time~\cite{chen1998atomic}.

\subsubsection{CRC}
In contrast with SRC, CRC finds the coefficient vector by solving the $\ell_2$-norm minimisation problem:
\begin{align}
	\label{equCRC}
	&\min{\parallel\boldsymbol{\alpha} \parallel}_2 \\
	&\text{s.t.    } \hspace{0.2cm} \mathbf{y}  =  \mathbf{X}\boldsymbol{\alpha}. \nonumber
\end{align}
The optimisation of Eq.~(\ref{equCRC}) is a typical least-square problem and $\boldsymbol{\alpha}$ can be obtained by:
\begin{equation}
	\label{equ4}
	\boldsymbol{\alpha} = (\mathbf{X}^{T}\mathbf{X}+\mu\mathbf{I})^{-1}\mathbf{X}^{T}\mathbf{y},
\end{equation}
where $ \mu $  is a small positive constant and $ \mathbf{I} $ is the identity matrix regularising the solution.
It has been shown that in certain conditions the $\ell_2$-norm based CRC offers competitive face classification accuracy as compared to the $\ell_1$-norm constrained SRC, and has much lower computational complexity~\cite{zhang2012collaborative}.
{\color{black}We propose a method that creates these conditions to enhance the performance of the CRC based face recognition.}

\section{The proposed method}
As discussed in Section \ref{section_1}, the problem of existing virtual-sample-generation algorithms is that they build on the intrinsic properties of a dataset, and are unable to cater for all possible appearance variations of a subject, \textit{i.e.} they are unable to inject new properties into an existing dictionary.
The problem of variations in appearance can only be mitigated using an over-complete dictionary that contains training samples {\color{black} covering the full spectrum of appearance variations}.
This motivates the search for better methods to capture full gamut of appearance variations by synthesising a set of virtual trainings samples using a 3D morphable face model for CRC-based face classification.

\subsection{Synthesising virtual samples with 3DMM}
A 3DMM is ideal for generating training samples with pose and illumination variations, and its use for this purpose is the tenet of our proposed method.
The 3DMM approach can reconstruct the 3D shape and texture of a 2D face image by fitting a generative 3D face model to the image.
To initialise the fitting process of our 3DMM, an automatic cascaded-regression-based facial landmark detection method is used~\cite{feng2015random}.
Then the reconstructed 3D shape and texture are used to render 2D face images with different poses by adjusting the parameters of a camera model.
For details of the 3DMM fitting algorithms the reader is referred to \cite{Tena_3DMM},~\cite{ME_ICB2013},~\cite{huber2015fitting} and~\cite{kittler20163d}, respectively.

\begin{figure}[t]
	\begin{center}
		\includegraphics[trim =20mm 40mm 20mm 15mm, clip, width=1 \linewidth]{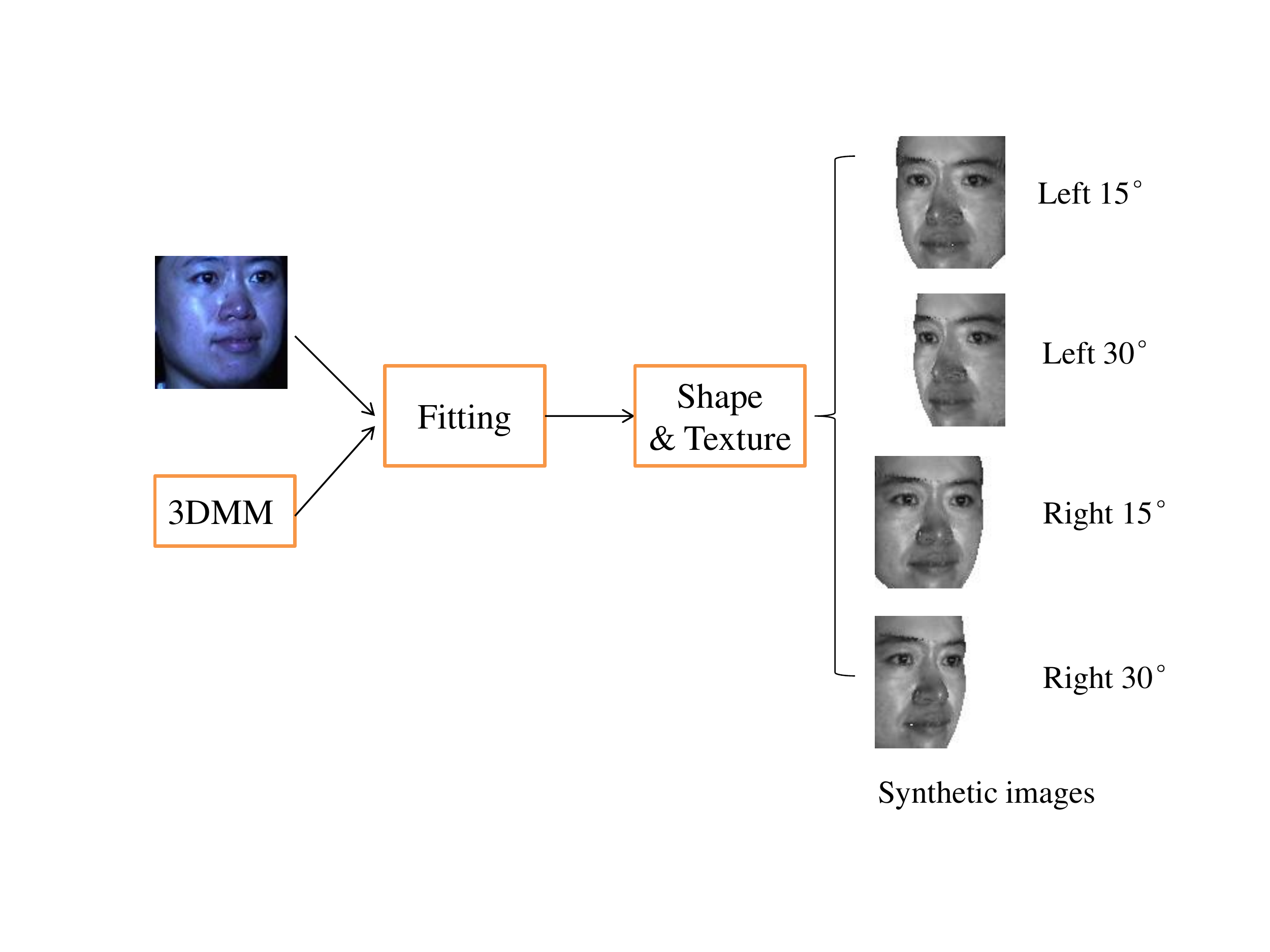}
	\end{center}
	\caption{Some rendered 2D faces from an input 2D image using 3DMM}
	\label{fig:sythesis}
\end{figure}

We render 2D virtual faces by projecting the reconstructed 3D shape and texture into a 2D image plane, using a perspective camera. 
More specifically, a vertex $ \mathbf{v} = [x^{3d}, y^{3d}, z^{3d}]^T \in \mathbb{R}^3$ of a 3D shape is projected to a 
2D coordinate $ \mathbf{s} = [x^{2d},y^{2d}]^T$ via a camera projection.
The projection can be decomposed into two parts: a rigid 3D transformation $T_r: \mathbb{R}^3 \rightarrow \mathbb{R}^3$ and a perspective projection $T_p: \mathbb{R}^3 \rightarrow \mathbb{R}^2$:
\begin{align} \label{GT1}
	T_r&:  \mathbf{v}' = \mathbf{R} \mathbf{v}+\boldsymbol{\tau}, \\
	T_p: \mathbf{s} &= \left[ \begin{array}{c} o_x+f \frac{v'_{x}} {v'_{z}} \\ o_y-f \frac{v'_{y}} {v'_{z}} \end{array} \right],
\end{align}
where $ \mathbf{R} \in \mathbb{R}^{3\times3}$ is the rotation matrix,
$\boldsymbol{\tau} \in \mathbb{R}^3$ is a spatial
translation, $f$ denotes the focal length, and $[o_{x},o_{y}]^T$ is the optical axis of the camera in the image plane.
Therefore, by setting different camera parameters  $\{\mathbf{R}, \boldsymbol{\tau},f\}$, images of different poses can be rendered 
from the reconstructed 3D shape and texture.
Some 2D face images rendered from an input face image using 3DMM are shown in Fig~\ref{fig:sythesis}.

\subsection{Exploiting representative samples from the extended dictionary}
To perform dictionary integration, we use the original and synthesised virtual faces to form an extended dictionary.
However, this extended training dataset consisting of virtual faces with different poses is redundant and may lead to inaccurate decision making.
In addition, due to the use of $\ell_2$-norm constraint, a CRC-based method cannot guarantee the sparsity of a reconstruction coefficient vector.
We therefore use the extended dictionary as an initial dictionary to be refined in the next step.
In order to decrease the adverse effects caused by improper hybrid training samples in CRC,
we use an elimination scheme to identify representative samples with the best capacity to represent a new sample. 

More specifically, we propose an iterative elimination scheme for discarding useless samples in the extended dictionary for face classification.
To this end, the contribution of each class to representing a test sample is measured in terms of reconstruction error.
Then all the training samples of the class with the largest reconstruction error are eliminated from the extended dictionary.
The coefficient vector of the extended dictionary and the contributions of the remaining classes are then updated.
The same process is repeated until the number of classes in the dictionary drops to a predefined level.

This elimination strategy strengthens those classes that are more informative and representative in reconstructing a test sample.  
In fact, we use Eq.~(\ref{equ_reconstuction_error}) to estimate the reconstruction error between a specific class and a test sample,
which is a distance measurement between a test sample and the linear combination of all training samples from the class.
A larger value of the reconstruction error means that the training samples of the class make tiny contributions in representing a test sample, and consequently this class should be eliminated from the extended dictionary. 
A further analysis to the proposed method is presented in the next section. 
The pipeline of our 3DPD-CRC face classification algorithm is shown in Fig.~\ref{Algorithm_3DPDCRC}.
\begin{figure}
\begin{algorithmic}[1]
\hrule
\vspace{0.05in}
\STATE \textbf{input} A dictionary consisting a set of training samples $ \mathbf{X} = [\mathbf{x}_{1,1},...,\mathbf{x}_{K,M}]$ and a test sample $\mathbf{y}$;

\STATE \textbf{preprocessing}: A 3DMM is used to perform 3D face reconstruction of $\mathbf{X}$ and to render a set of virtual faces
$\hat{\mathbf{X}} = [\hat{\mathbf{x}}_{1,1},..., \hat{\mathbf{x}}_{K,V}]$ that are used as an auxiliary dataset to form the extended dictionary $\tilde{\mathbf{X}} = [\mathbf{X}, \hat{\mathbf{X}}]$;

\FOR{$l = 1$ to $L$ (a pre-defined parameter)}
\STATE Encode the test sample using CRC and obtain the coefficient vector, as described in Eq.~(\ref{equ4});

\STATE Compute the reconstruction error of each class using Eq.~(\ref{equ_reconstuction_error}) and 
eliminate all the training samples of the class achieving the largest reconstruction error to update the dictionary;

\ENDFOR

\RETURN The label of the test sample using Eq.~(\ref{equ11_2}).

\vspace{0.03in}
\hrule
\end{algorithmic}
\caption{The proposed 3DPD-CRC algorithm}
\label{Algorithm_3DPDCRC}
\end{figure}

\section{Analysis of the proposed method}
To reveal the nature of the proposed method, in this section, we further analyse our 3DPD-CRC from both theoretical and empirical perspectives. 

\subsection{Improvements and the underlying rationale}
{\color{black} 
Let $\tilde{\mathbf{X}}$ denote the augmented dictionary, created from the original training set $\mathbf{X}$ and the synthesised set $\hat{\mathbf{X}}$. 
Further, let $\widetilde{ \boldsymbol{\alpha}}_i = [\alpha_{i1},....,\alpha_{iM},\hat \alpha_{i1},.....,\hat \alpha_{iV}]^T$ be the vector of CRC coefficients reconstructing 
input pattern ${\bf y}$,  using the augmented dictionary ${\bf \tilde X}$. Let us assume that ${\bf y}$ belongs to class $i$. 

In order to explain the need for the proposed augmentation of the training set and the on-line dictionary pruning by hypothesis elimination, we shall consider a few examples: \\
{\bf Case 1:} Suppose the synthesised training samples are not available. 
Then only the coefficients for class $i$ associated with the original training samples are non-zero, 
i.e. $\widetilde{ \boldsymbol{\alpha}}_i = [\alpha_{i1},....,\alpha_{iM},0,....,0]^T$. 
However, as the data set does not contain enough samples to represent different poses, the $i^{th}$ class 
fitting error $||\mathbf{y}- {\bf \tilde X_i} \widetilde{ \boldsymbol{\alpha}}_i||_2$ will be quite high, causing misclassification.\\
{\bf Case 2:} Suppose we have injected (by means of synthesised samples) dictionary items which represent sample ${\bf y}$ 
very well. This will be reflected in coefficients $\alpha_{ij}$ taking values close to zero, 
\textit{i.e.} $\widetilde{\boldsymbol{\alpha}}_i \approx [0,....,0,\hat \alpha_{i1},....,\hat \alpha_{iV}]^T$. 
However, if at the same time we have injected redundancy that is enabling samples from other classes 
to contribute actively to the reconstruction of pattern ${\bf y}$, this will create an opportunity 
for CRC to dilute the strength of coefficients $\hat \alpha_{ij}$ and distribute their weight over 
samples from the other classes, \textit{i.e.} over coefficients $\hat \alpha_{kj}, \forall k \neq i$. 
As these samples furnish similar information, their impact is that the total weight needed 
for the reconstruction is divided between them.
For the same approximation error, the $\ell_2$ norm minimisation will prefer this weight-diluting solution, as the sum of many small values squared is much smaller 
than the sum of a few larger weights squared. 
The reconstruction of ${\bf y}$ in the presence of 
redundancy will reduce the weights of samples from class $i$, increasing the approximation error, and potentially leading to misclassification. \\
{\bf Case 3:} If a systematic on-line elimination of the training samples  from the clutter hypotheses 
(classes with high approximation error) is carried out, the redundancy is suppressed. 
The pruning process will increase the weight of coefficients $\hat \alpha_{ij}$ and enhance 
their ability to reconstruct the input pattern with low error, thus leading to correct identification 
of the class membership of ${\bf y}$. The hypothesis elimination process induces sparsity in a manner 
similar to the Iterative Hard Thresholding algorithm~\cite{blumensath2009iterative}.
} 

\subsection{An empirical explanation of the proposed method}
\begin{figure*}[t]
	\centering
	\subfloat[The original dictionary]{
		\label{fig3_a}
		\includegraphics[trim = 30mm 22mm 40mm 15mm, clip, width=0.44\linewidth]{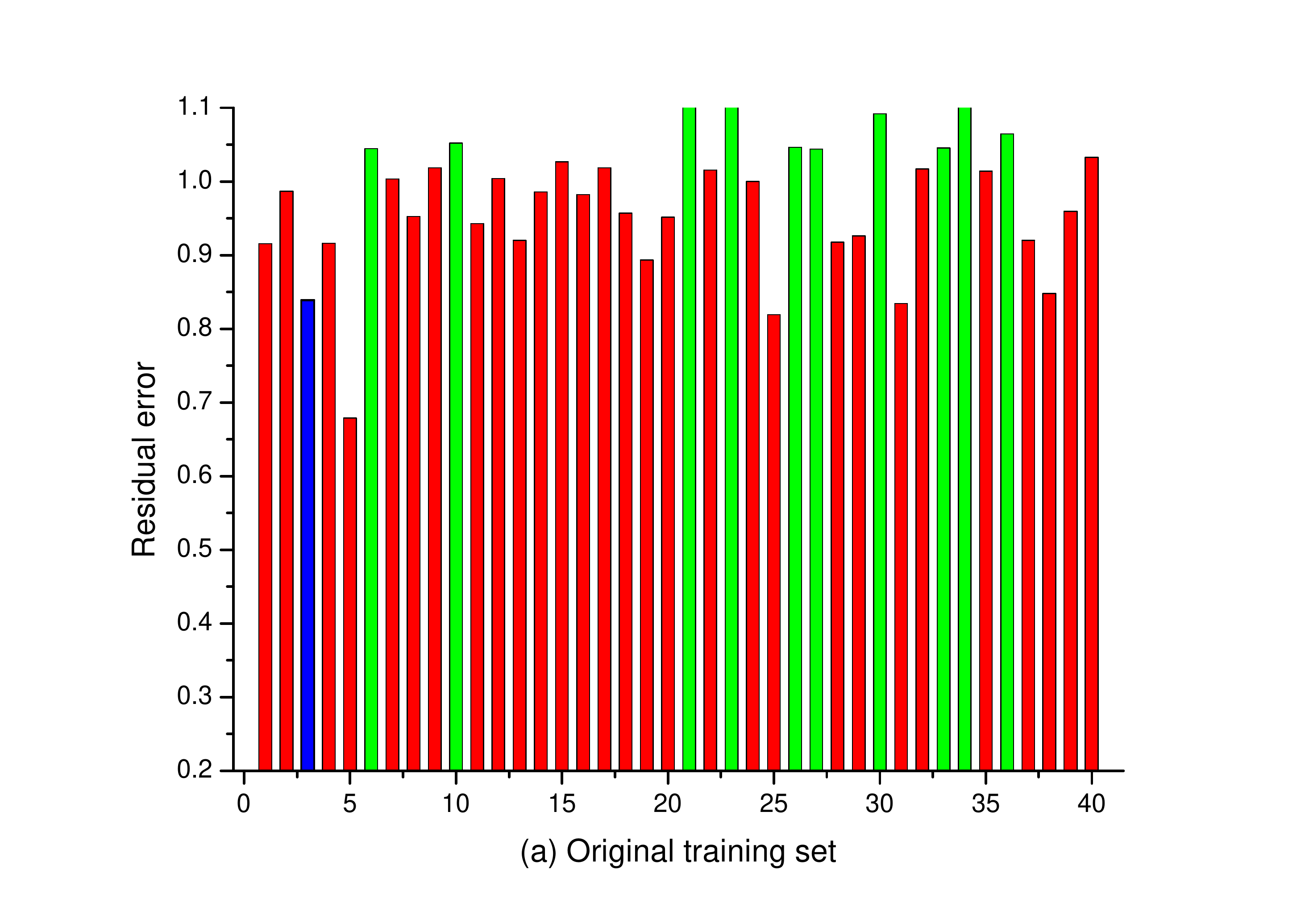}
	}
	\subfloat[The original dictionary with the elimination strategy]{
		\label{fig3_b}
		\includegraphics[trim = 30mm 22mm 40mm 15mm, clip, width=0.44\linewidth]{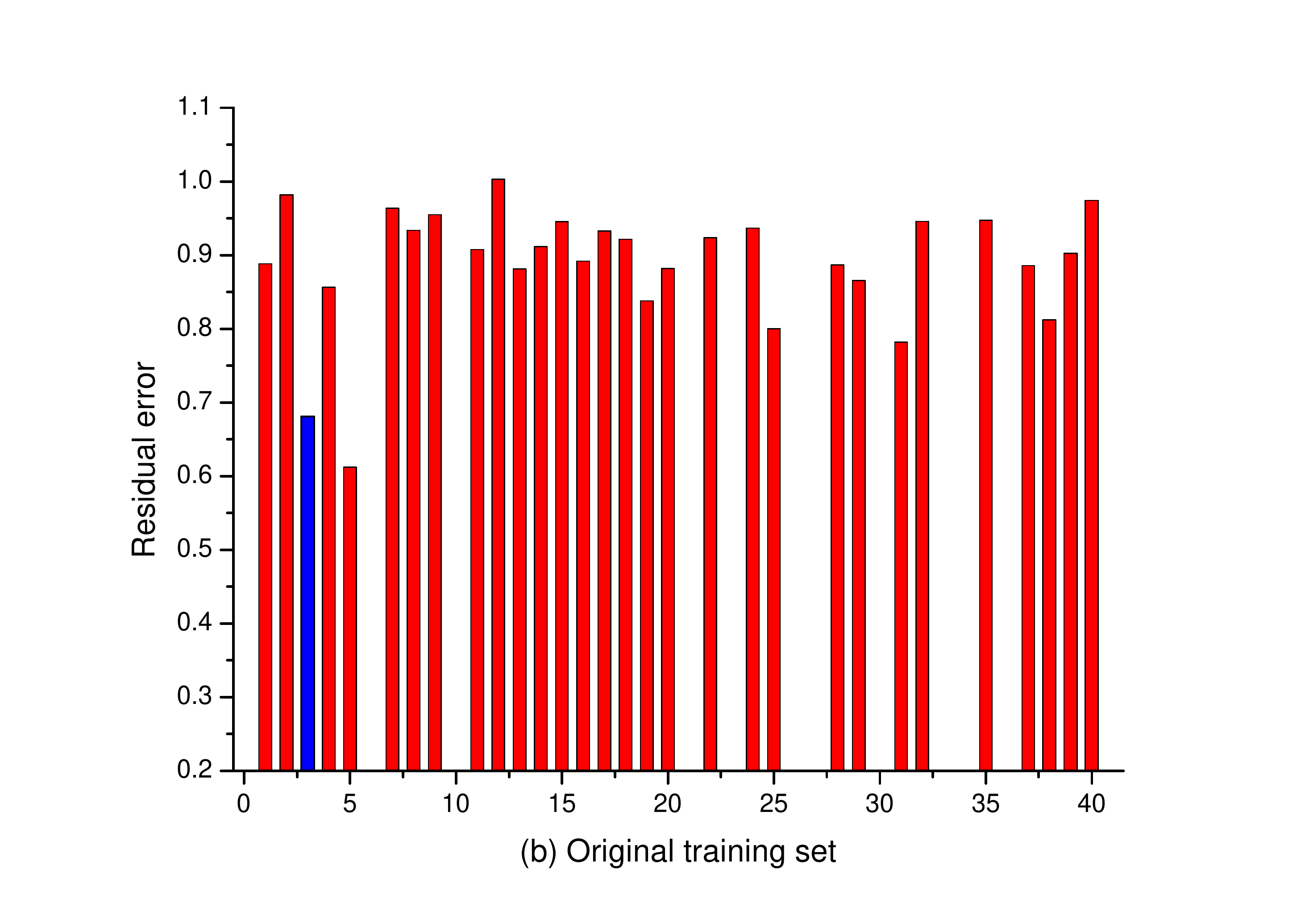}
	}
	
		\subfloat[The extended dictionary]{
		\label{fig3_c}
		\includegraphics[trim = 30mm 25mm 40mm 15mm, clip, width=0.44\linewidth]{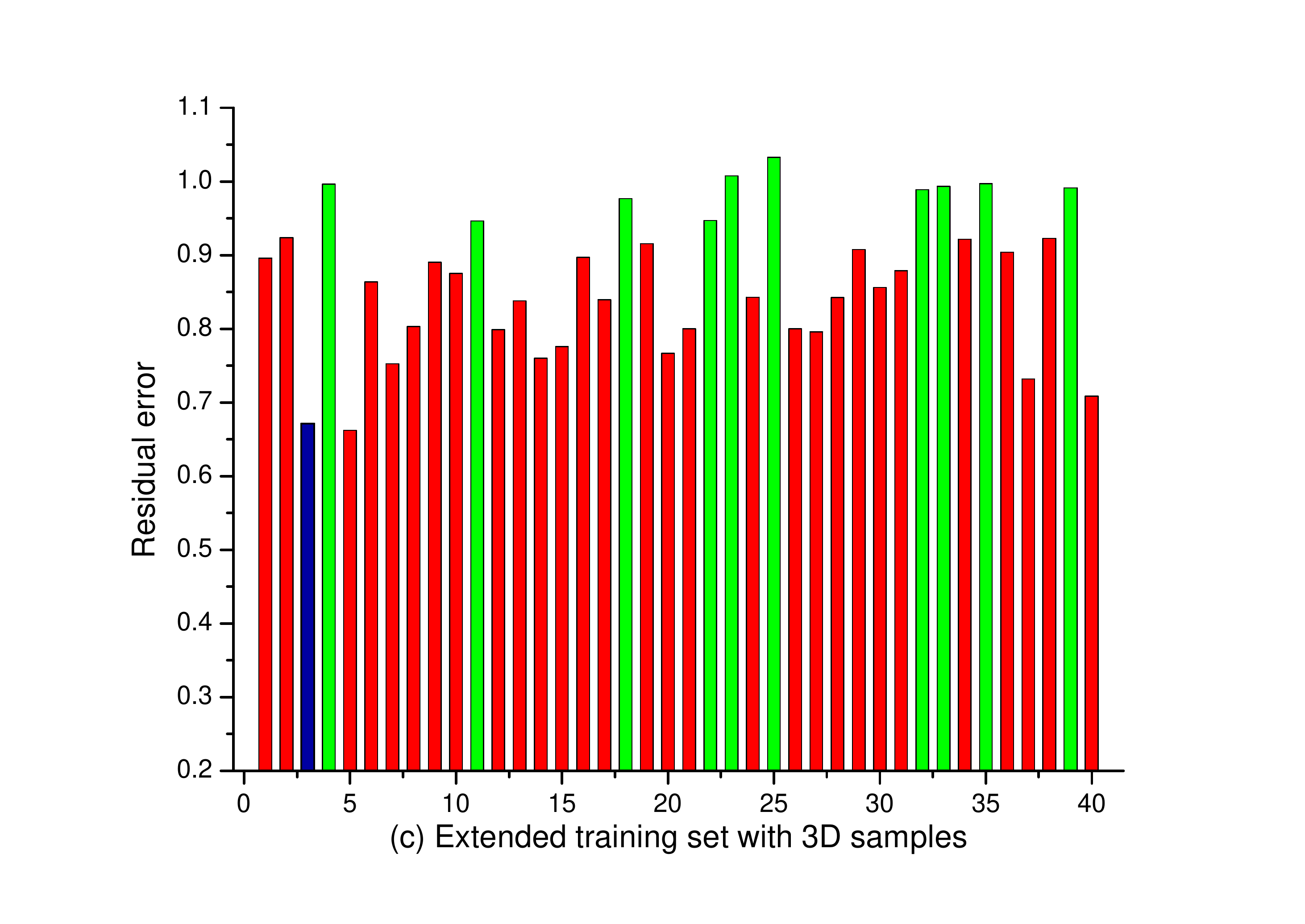}
	}
	\subfloat[The extended dictionary with the elimination strategy]{
		\label{fig3_d}
		\includegraphics[trim = 30mm 25mm 40mm 15mm, clip, width=0.44\linewidth]{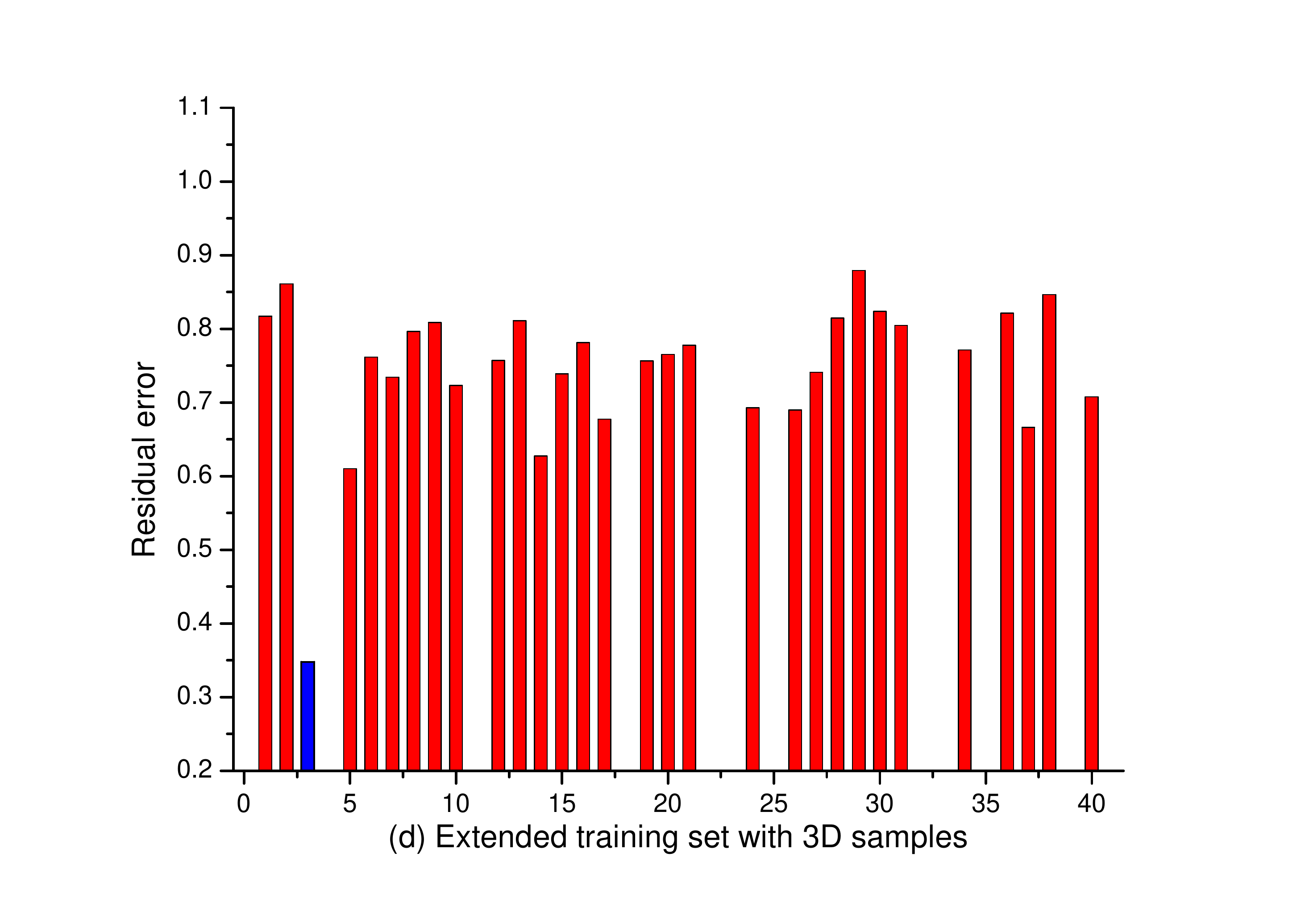}
	}
	\caption{Reconstruction errors of a test image using: (a) the original dictionary; (b) the original dictionary with the elimination strategy; (c) the extended dictionary consisting of virtual faces generated by 3DMM; and (d) the extended dictionary with the elimination strategy. The blue bars indicate the correct label for the test sample, and the green bars indicate the classes should be discarded from the dictionary using the proposed elimination scheme.}
	\label{fig3}
\end{figure*}

In this section, we present an empirical explanation {\color{black} of} the proposed 3DPD-CRC algorithm.
To demonstrate how the proposed method works, Fig.~\ref{fig3} shows the reconstruction error of a test sample using the training samples of each class in the dictionary, evaluated on the ORL face dataset that has 40 subjects and each subject has 10 face images.
We used the first two face images per subject as training samples and the remaining 8 images as test samples.
Fig.~\ref{fig3} shows the reconstruction errors of a randomly selected test sample from the 3rd subject.
The reconstruction errors of the test sample by the correct class are highlighted using blue bars. 
Green bars indicate the classes with higher reconstruction errors and should be discarded during the elimination scheme.

The reconstruction errors of the selected test sample using the classical CRC algorithm by 40 classes of the original dictionary without elimination are shown in Fig.~\ref{fig3}(a).
The 3rd class does not have the minimal reconstruction error and the label of the test sample is assigned to the 5th class that provides best representation to the test sample.
According to the underlying assumption of the proposed elimination scheme, a larger error indicates that the corresponding class (with green bar) in the dictionary has tiny effects {\color{black} on representing} a test sample hence should be eliminated from {\color{black} the} dictionary.
Hence, we iteratively {\color{black} discard} some classes from the original dictionary and {\color{black} re-calculate} the reconstruction error of the test sample by each class, as shown in Fig.~\ref{fig3}(b).
The reconstruction error of the test sample by the third class is reduced when using {\color{black} fewer} classes in the {\color{black} original} dictionary, but it {\color{black} is} still higher than that of the 5th class and lead to inaccurate decision making.

To {\color{black} demonstrate the merit} of the proposed data augmentation method, we repeated the above procedure using the extended dictionary with 3DMM synthesised virtual training faces.
The results are shown in Fig.~\ref{fig3}(c) (without elimination) and Fig.~\ref{fig3}(d) (with elimination).
The single use of the extended dictionary also reduces the reconstruction error of the test sample by the correct class, \textit{i.e.} the 3rd class, as shown in Fig.~\ref{fig3}(c).
However, the reconstruction error of the 5th class is still the minimal one thereby leading to an incorrect face classification result.
But, as shown in Fig.~\ref{fig3}(d), the reconstruction error of the test sample from the 3rd class is greatly reduced and the correct classification result can be achieved by jointly using the extended dictionary and the elimination scheme.

From this experiment, we can suggest that the joint use of virtual training samples and the elimination scheme in our 3DPD-CRC improves the accuracy of face classification.
Moreover, the proposed method results in a dictionary learned from a dynamic optimisation process, which increases the sparsity of the reconstruction coefficient vectors obtained by CRC.

\section{Experimental Results}
\label{section_experiments}
In this section, we evaluate the proposed 3DPD-CRC algorithm on three face datasets: ORL~\cite{heap1995real}, FERET~\cite{phillips2000feret} and PIE~\cite{sim2003cmu}. 

\begin{figure}[t]
	\centering
	\subfloat[ORL]{
	\label{fig4_1}
		\includegraphics[width=1\linewidth]{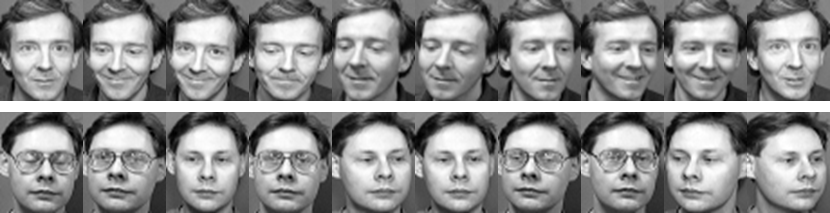}
		}
		
	\subfloat[FERET]{
	\label{fig4_2}
		\includegraphics[width=.7\linewidth]{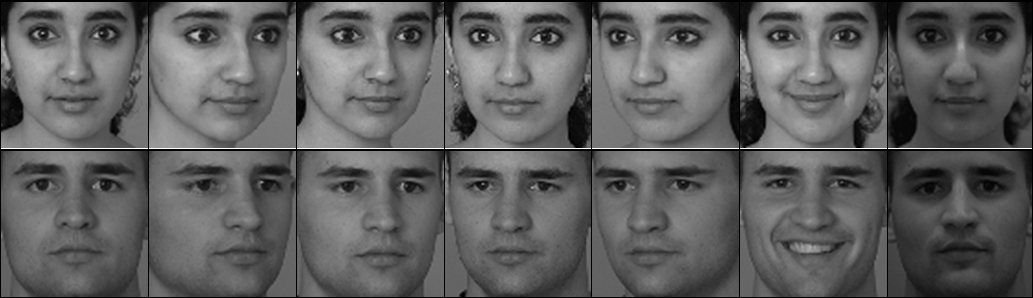}
		}
		
	\subfloat[PIE]{
	\label{fig4_3}
	\includegraphics[width=.5\linewidth]{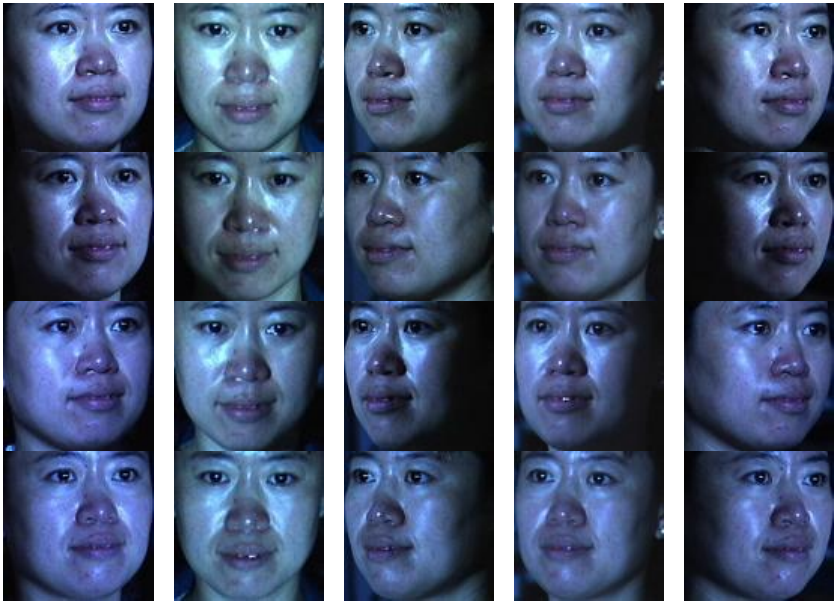}
	}
	\caption{Example faces of the ORL, FERET and PIE datasets}
	\label{fig4}
\end{figure}

The ORL dataset contains 40 subjects and each subject has 10 face images.
The images were captured at different time instances, with slightly varying lighting conditions, expressions, and artefacts.
Some examples of ORL are shown in Fig.~\ref{fig4_1}.

The FERET dataset is a result of the FERET program, which was sponsored by the U.S. Department of Defence through the DARPA program~\cite{phillips2000feret}.
It has become a very popular benchmarking dataset for the evaluation of face recognition techniques.
The proposed algorithm was evaluated on a subset of FERET, which includes 1400 images of 200 individuals with 7 different images per subject.
Some examples of the FERET dataset are shown in Fig.~\ref{fig4_2}.

The CMU PIE dataset consists of 41,368 images of 68 individuals with mixed variations in pose, expression and illumination.
The images of each subject were captured under 13 poses, 43 illuminations and 4 expressions.
The proposed algorithm was evaluated on a subset of the PIE dataset, which includes 1360 images of 68 subjects.
Each subject has 5 pose variations and 4 illumination variations, as shown in Fig.~\ref{fig4_3}.

\subsection{Results on ORL}
For the ORL face dataset, {\color{black}we followed the evaluation protocol that has been widely used in previous studies~\cite{xu2011two, deng2012extended, xu2013using1}.}
We randomly selected $ \theta (\theta=2,3,4) $ samples of each subject for training and the remaining ones were used for test.
Thus, a training set of $ 40\times \theta $ images and a test set with $ 40\times(10-\theta) $ images were created in each experiment.
We repeated our experiment 10 times and measured the accuracy of different face classification algorithms in terms of recognition rate.
Meanwhile, we applied 3DMM fitting to each training sample and synthesised 10 virtual faces with $\pm4^\circ$, $\pm8^\circ$, $\pm12^\circ$, $\pm16^\circ$ and $\pm20^\circ$ yaw rotations.
The elimination scheme presented in Section 3.2 was performed in classification. 

The classification results of SRC~\cite{wright2009robust}, CRC~\cite{zhang2012collaborative} 
and the proposed 3DPD-CRC on ORL with 2, 3 and 4 training samples are presented in Table~\ref{tab1}, Table~\ref{tab2} and Table~\ref{tab3}, respectively.
In these tables, the term `elimination proportion' indicates the proportion of the removed classes in the elimination phase.
It should be noted that the elimination strategy was used for all these three algorithms. 
As shown in Table~\ref{tab1}, \ref{tab2} and \ref{tab3}, the proposed 3DPD-CRC method using the extended hybrid dictionary outperforms the classical CRC and SRC in terms of accuracy, regardless {\color{black} of} the proportion of the eliminated classes and the number of training samples.
The results validate the effectiveness of the proposed method {\color{black} of} jointly using synthesised virtual faces and the elimination scheme.
However, it is hard to determine the best value of the elimination proportion because different methods perform best at different proportions of the eliminated classes.
One practical solution to this issue is to tune this parameter using cross validation for a specific face recognition task.
\begin{table*}[!t]
\renewcommand{\arraystretch}{1.2}
	\footnotesize
	\centering
	\caption{Face recognition rates (\%) of different methods with 2 randomly selected training samples on ORL}
	\begin{tabular}{cclcccccccc}
		\hline
		\multirow{2}{*}{Method} & \multicolumn{10}{c}{Elimination Proportion}                                                                                            \\
		\cline{2-11}
		& \multicolumn{2}{c}{10\%}      & 20\%      & 30\%      & 40\%      
		& 50\%      & 60\%      & 70\%      & 80\%      & 90\%     \\
		\hline          
		3DPD-CRC                & \multicolumn{2}{c}{\textbf{85.5$\pm$0.06}} & \textbf{86.3$\pm$0.05} & \textbf{86.3$\pm$0.05} 
		& \textbf{86.7$\pm$0.06} & \textbf{87.0$\pm$0.08} & \textbf{87.0$\pm$0.05} & \textbf{87.2$\pm$0.07} & \textbf{87.3$\pm$0.07} & \textbf{\textit{88.0$\pm$0.05}} \\
		
		SRC                     & \multicolumn{2}{c}{52.2$\pm$2.99}  & 79.2$\pm$1.53  & 83.7$\pm$2.24  
		& 85.4$\pm$2.51  & 85.4$\pm$2.47  &   84.7$\pm$2.26  & 85.7$\pm$2.29 & 83.7$\pm$2.32  & 82.6$\pm$2.07 \\
		
		CRC                     & \multicolumn{2}{c}{82.8$\pm$2.48} & 83.4$\pm$2.37 & 83.9$\pm$2.16 
		& 84.5$\pm$2.62 & 85.4$\pm$2.65 &  85.9$\pm$1.93 & 86.2$\pm$2.12  &  85.6$\pm$2.49  & 84.1$\pm$2.47   \\        
		\hline 
	\end{tabular}
	\label{tab1}
\end{table*}
\begin{table*}[!t]
\renewcommand{\arraystretch}{1.2}
	\footnotesize
	\centering
	\caption{Face recognition rates (\%)  of different methods with 3 randomly selected training samples on ORL}
	\begin{tabular}{cclcccccccc}
		\hline
		\multirow{2}{*}{Method} & \multicolumn{10}{c}{Elimination Proportion}                                                                                            \\
		\cline{2-11}
		& \multicolumn{2}{c}{10\%}      & 20\%      & 30\%     
		& 40\%      & 50\%      & 60\%      & 70\%      & 80\%      & 90\%     \\
		\hline          
		3DPD-CRC  & \multicolumn{2}{c}{\textbf{91.6$\pm$0.11}} & \textbf{91.3$\pm$0.08} & \textbf{92.5$\pm$0.11} 
		& \textbf{92.5$\pm$0.11} & \textbf{\textit{93.4$\pm$0.14}} & \textbf{93.1$\pm$0.12} & \textbf{93.1$\pm$0.11} & \textbf{92.8$\pm$0.11} & \textbf{92.8$\pm$0.12} \\
		
		SRC       & \multicolumn{2}{c}{79.5$\pm$1.90} & 89.4$\pm$1.66 & 90.0$\pm$1.37 
		& 91.0$\pm$1.17 & 90.5$\pm$1.42 & 89.8$\pm$1.54 & 91.1$\pm$0.95 & 89.0$\pm$1.70 & 88.7$\pm$1.72  \\
		
		CRC       & \multicolumn{2}{c}{88.1$\pm$1.79} & 88.3$\pm$2.08 & 89.0$\pm$1.76 
		& 89.1$\pm$1.59 & 90.3$\pm$1.64 & 91.2$\pm$1.21 & 91.3$\pm$1.35 &  91.6$\pm$0.89 &  90.6$\pm$1.60    \\
		\hline 
	\end{tabular}
	\label{tab2}
\end{table*}
\begin{table*}[!t]
\renewcommand{\arraystretch}{1.2}
	\footnotesize
	\centering
	\caption{Face recognition rates (\%) of different methods with 4 randomly selected training samples on ORL}
	\begin{tabular}{cclcccccccc}
		\hline
		\multirow{2}{*}{Method} & \multicolumn{10}{c}{Elimination Proportion}                                                                                            \\
		\cline{2-11}
		& \multicolumn{2}{c}{10\%}      & 20\%      & 30\%      
		& 40\%      & 50\%      & 60\%      & 70\%      & 80\%      & 90\%     \\
		\hline          
		3DPD-CRC                & \multicolumn{2}{c}{\textbf{96.5$\pm$0.02}} & \textbf{96.9$\pm$0.02} 
		& \textbf{97.1$\pm$0.02} & \textbf{97.1$\pm$0.02} & \textbf{96.9$\pm$0.02} & \textbf{96.8$\pm$0.02} & \textbf{96.8$\pm$0.02} 
		& \textbf{96.9$\pm$0.02} & \textbf{\textit{97.2$\pm$0.02}} \\
		
		SRC                     & \multicolumn{2}{c}{91.1$\pm$1.26} & 92.3$\pm$1.68 
		& 92.9$\pm$1.51 & 92.8$\pm$1.25 & 92.2$\pm$1.20 &           92.2$\pm$1.24 & 93.7$\pm$1.30 & 91.8$\pm$1.80 & 91.0$\pm$1.39 \\
		
		CRC                     & \multicolumn{2}{c}{90.1$\pm$1.60} & 91.0$\pm$1.87 
		& 91.4$\pm$1.89 & 91.9$\pm$1.52 & 92.0$\pm$1.51 &           92.5$\pm$1.26 & 93.0$\pm$1.05  &  93.8$\pm$0.94  & 92.7$\pm$0.93   \\
		\hline 
	\end{tabular}
	\label{tab3}
\end{table*}

Table~\ref{tab4} presents the recognition rates achieved by a set of traditional face classification methods including SRC~\cite{wright2009robust}, CRC~\cite{zhang2012collaborative}, LRC~\cite{naseem2010linear}, 
L21SDA~\cite{shi2014face}, TPTSR~\cite{xu2011two}, ESRC~\cite{deng2012extended}, CFFR~\cite{xu2013using1} and SFRC~\cite{xu2013using}, as well as the proposed 3DPD-CRC method, using 2 or 3 training samples of each class in the original dictionary. 
The proposed 3DPD-CRC method achieves 88.0$\%$ and 92.8$\%$ recognition rates when using only 2 and 3 samples per subject as training samples. 
These results are better than those achieved by all the other methods.
\begin{table}[t]
\renewcommand{\arraystretch}{1.2}
	\centering
	\caption{Face recognition rates $(\%)$ of different methods on ORL}
	\begin{tabular}{cclc}
		\hline
		\multirow{2}{*}{Method} & \multicolumn{3}{c}{\begin{tabular}[c]{@{}c@{}}Number of training samples\end{tabular}} \\
		\cline{2-4}
		& \multicolumn{2}{c}{2}                             & 3                                    \\
		\hline
		SRC                     & \multicolumn{2}{c}{85.7}                          & 91.1                                 \\
		CRC                     & \multicolumn{2}{c}{86.2}                          & 91.6                                 \\
		LRC                     & \multicolumn{2}{c}{84.6}                          & 90.2                                 \\
		SDA-L2                  & \multicolumn{2}{c}{80.5}                          & 82.1                                 \\
		TPTSR                   & \multicolumn{2}{c}{83.4}                          & 87.8                                 \\
		ESRC                    & \multicolumn{2}{c}{87.1}                          & 89.6                                 \\
		CFFR                    & \multicolumn{2}{c}{83.2}                          & 88.4                                 \\
		SFRC                    & \multicolumn{2}{c}{87.7}                          & 91.3                                 \\
		3DPD-CRC                & \multicolumn{2}{c}{\textbf{88.0}}                 & \textbf{92.8}                        \\
		\hline                   
	\end{tabular}
	\label{tab4}
\end{table}

\subsection{Results on FERET}
For the FERET dataset, the same procedure as in ORL was used to split the original dataset into training and test sets.
{\color{black} This evaluation protocol is compliant with that used in similar experiments reported in the literature.}
The number of training samples per subject was set to $\theta (\theta=2,3,4)$, which resulted in a training set with $ 200\times \theta $ images and a test set with $ 200\times(7-\theta) $ images.
To obtain the extended dictionary, we used 3DMM to fit each training sample and rendered 10 virtual samples with the same pose variations as in the last section.

The face classification results of SRC, CRC, and our 3DPD-CRC on FERET are shown in Table~\ref{tab5}, Table~\ref{tab6} and Table~\ref{tab7} using 2, 3 and 4 training samples per subject in the original dictionary.
The elimination strategy was used for all these three methods.
As shown in these tables, in {\color{black} conjunction} with the elimination scheme, the proposed 3DPD-CRC method consistently achieves better classification results than SRC and CRC, regardless of the elimination propotion and the number of training samples.
\begin{table*}[t]
\renewcommand{\arraystretch}{1.2}
	\scriptsize
	\centering
	\caption{Face recognition rates $(\%)$ of different methods with 2 randomly selected training samples on FERET}
	\begin{tabular}{cclcccccccc}
		\hline
		\multirow{2}{*}{Method} & \multicolumn{10}{c}{Elimination Proportion}                                                                                            \\
		\cline{2-11}
		& \multicolumn{2}{c}{10\%}      & 20\%      & 30\%      
		& 40\%      & 50\%      & 60\%      & 70\%      & 80\%      & 90\%     \\
		\hline          
		3DPD-CRC  & \multicolumn{2}{c}{\textbf{77.7$\pm$0.36}} & \textbf{79.4$\pm$0.34} 
		& \textbf{80.6$\pm$0.29} & \textbf{81.1$\pm$0.25} & \textbf{81.1$\pm$0.26} & \textbf{\textit{81.5$\pm$0.17}} & \textbf{81.4$\pm$0.15} & \textbf{81.3$\pm$0.16} & \textbf{81.0$\pm$0.19} \\
		
		SRC   & \multicolumn{2}{c}{48.6$\pm$12.27} & 49.5$\pm$12.14 
		& 50.1$\pm$11.88 & 50.7$\pm$11.77 & 51.7$\pm$11.87 & 52.5$\pm$11.74  &  53.3$\pm$11.32 &  54.1$\pm$11.15  & 55.4$\pm$10.67   \\
		
		CRC  & \multicolumn{2}{c}{45.7$\pm$10.08}   & 46.6$\pm$10.27  
		& 47.9$\pm$9.95 &  48.9$\pm$ 10.03 & 50.6$\pm$9.69 & 52.2$\pm$10.08   & 54.0$\pm$10.0  & 55.1$\pm$10.14  & 55.6$\pm$9.96   \\
		
		\hline 
	\end{tabular}
	\label{tab5}
\end{table*}
\begin{table*}[t]
\renewcommand{\arraystretch}{1.2}
	\scriptsize
	\centering
	\caption{Face recognition rates $(\%)$ of different methods with 3 randomly selected training samples on FERET}
	\begin{tabular}{cclcccccccc}
		\hline
		\multirow{2}{*}{Method} & \multicolumn{10}{c}{Elimination Proportion}     \\
		\cline{2-11}
		& \multicolumn{2}{c}{10\%}      & 20\%      & 30\%      
		& 40\%      & 50\%      & 60\%      & 70\%      & 80\%      & 90\%     \\
		\hline          
		3DPD-CRC                & \multicolumn{2}{c}{\textbf{93.0$\pm$0.34}} & \textbf{93.6$\pm$0.26} 
		& \textbf{\textit{94.0$\pm$0.24}} & \textbf{93.5$\pm$0.25} & \textbf{93.5$\pm$0.26} & \textbf{93.4$\pm$0.19} & \textbf{93.4$\pm$0.22} & \textbf{93.0$\pm$0.22} & \textbf{92.6$\pm$0.19}
		\\
		SRC                     & \multicolumn{2}{c}{65.7$\pm$10.33}   & 65.7$\pm$10.08 
		& 65.8$\pm$10.31 & 66.2$\pm$10.21 & 66.5$\pm$10.44 & 67.0$\pm$10.38 & 67.2$\pm$10.22 & 67.5$\pm$10.32 & 68.0$\pm$10.28\\
		CRC                     & \multicolumn{2}{c}{58.7$\pm$9.82} & 59.6$\pm$9.99  
		& 60.6$\pm$10.18 & 61.7$\pm$10.31 & 62.7$\pm$9.91 & 64.3$\pm$10.56  & 65.6$\pm$10.14  & 67.8$\pm$10.08  & 68.7$\pm$9.45\\
		\hline 
	\end{tabular}
	\label{tab6}
\end{table*}
\begin{table*}[t]
\renewcommand{\arraystretch}{1.2}
	\scriptsize
	\centering
	\caption{Face recognition rates $(\%)$ of different methods with 4 randomly selected training samples on FERET}
	\begin{tabular}{cclcccccccc}
		\hline
		\multirow{2}{*}{Method} & \multicolumn{10}{c}{Elimination Proportion}                                                                                            \\
		\cline{2-11}
		& \multicolumn{2}{c}{10\%}      & 20\%      & 30\%      
		& 40\%      & 50\%      & 60\%      & 70\%      & 80\%      & 90\%     \\
		\hline          
		3DPD-CRC                & \multicolumn{2}{c}{\textbf{96.5$\pm$0.29}} & \textbf{96.9$\pm$0.25} 
		& \textbf{\textit{97.1$\pm$0.25}} & \textbf{97.1$\pm$0.27} & \textbf{96.9$\pm$0.19} & \textbf{96.8$\pm$0.20} & \textbf{96.8$\pm$0.19} & \textbf{96.9$\pm$0.17} & \textbf{96.5$\pm$0.20}
		\\
		SRC                     & \multicolumn{2}{c}{72.0$\pm$13.49}  & 72.1$\pm$13.51 
		& 72.3$\pm$13.35  & 72.2$\pm$13.42 & 73.1$\pm$13.91  & 73.8$\pm$13.74 & 74.1$\pm$13.48 & 74.4$\pm$13.38  & 74.8$\pm$12.45\\
		CRC                     & \multicolumn{2}{c}{62.2$\pm$12.68}  & 62.9$\pm$12.73 
		& 64.5$\pm$13.13 & 65.7$\pm$13.90 & 67.5$\pm$13.50  & 68.9$\pm$13.65 & 70.3$\pm$13.43 & 72.5$\pm$13.10  & 74.5$\pm$11.93\\
		\hline 
	\end{tabular}
	\label{tab7}
\end{table*}

The face classification results of SRC~\cite{wright2009robust}, CRC~\cite{zhang2012collaborative}, 
LRC~\cite{naseem2010linear}, L21SDA~\cite{shi2014face}, TPTSR~\cite{xu2011two}, ESRC~\cite{deng2012extended}, 
CFFR~\cite{xu2013using1}, SFRC~\cite{xu2013using}, PCA+LDA~\cite{yang2003can}  and our 3DPD-CRC on the FERET dataset are presented in Table~\ref{tab8}.
The table presents the face recognition rates of different algorithms using both 2 and 3 randomly selected training samples per class in the original dictionary.
We repeated our experiment 10 times and {\color{black}report} the average recognition rate.
According to this table, the proposed 3DPD-CRC method achieves much better results than other methods in terms of recognition rate.
\begin{table}[t]
\renewcommand{\arraystretch}{1.2}
	\centering
	\caption{Face recognition rates $(\%)$ of different methods on FERET}
	\begin{tabular}{lcc}
		\hline
		\multirow{2}{*}{Method} & \multicolumn{2}{c}{Number of training samples} \\
		\cline{2-3}
		& 2                     & 3               \\
		\hline
		SRC           & 55.4                  & 68.0            \\
		CRC           & 55.6                  & 68.7            \\
		LRC           & 66.0                  & 74.0            \\
		TPTSR         & 59.9                  & 68.7            \\
		ESRC          & 58.7                  & 69.5            \\
		CFFR          & 56.4                  & 66.8            \\
		SFRC          & 67.9                  & 74.2            \\
		PCA+LDA       & 52.5                  & 62.6            \\
		3DPD-CRC      & \textbf{81.5}         & \textbf{94.0}   \\
		\hline                   
	\end{tabular}
	\label{tab8}
\end{table}

\subsection{Results on PIE}
\begin{figure*}[t]
	\centering
	\subfloat[2 training samples]{
		\label{fig7_a}
		\includegraphics[trim = 36mm 95mm 42mm 100mm, clip, width=0.32\linewidth]{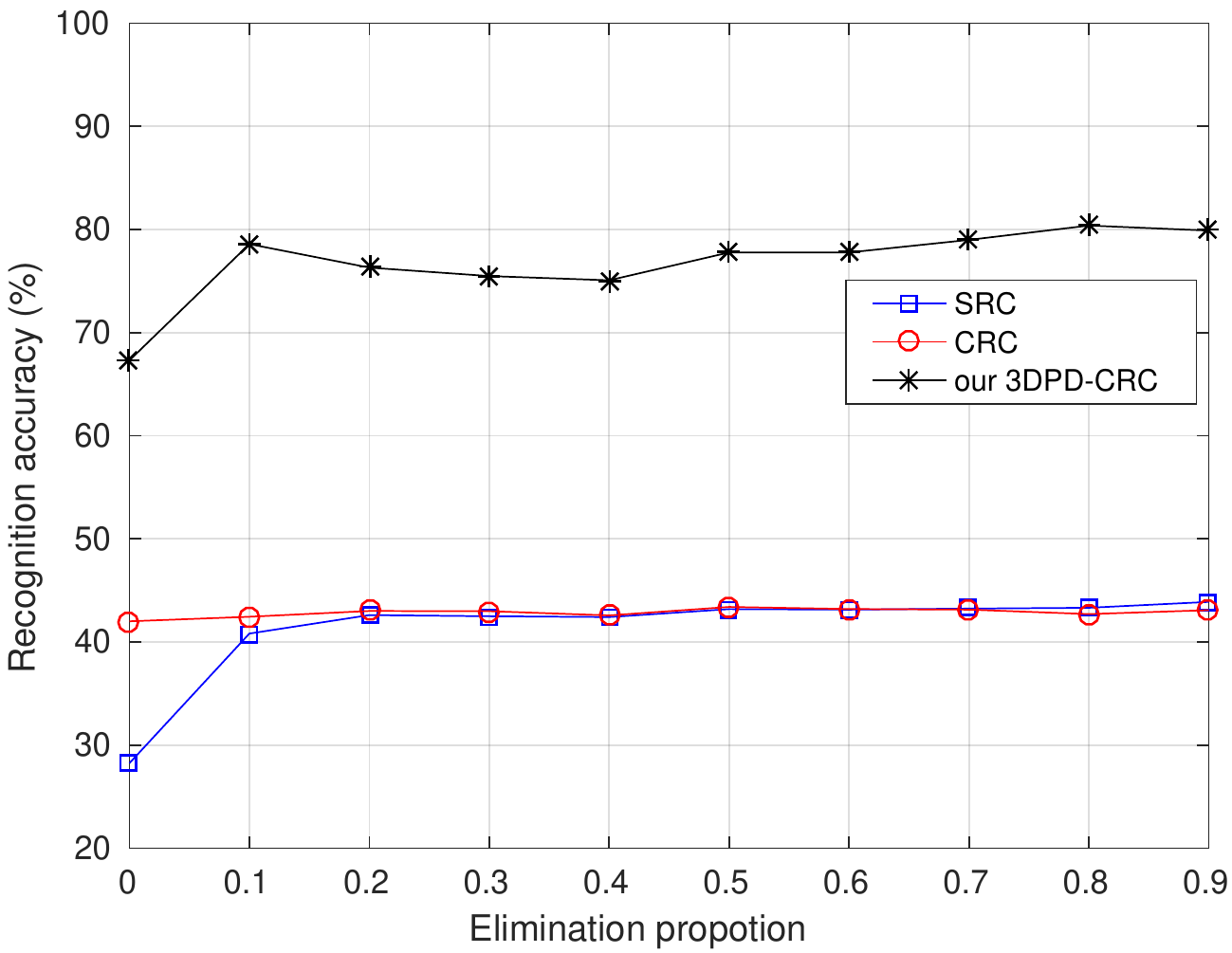}
	}
	\subfloat[3 training samples]{
		\label{fig7_b}
		\includegraphics[trim = 36mm 95mm 42mm 100mm, clip, width=0.32\linewidth]{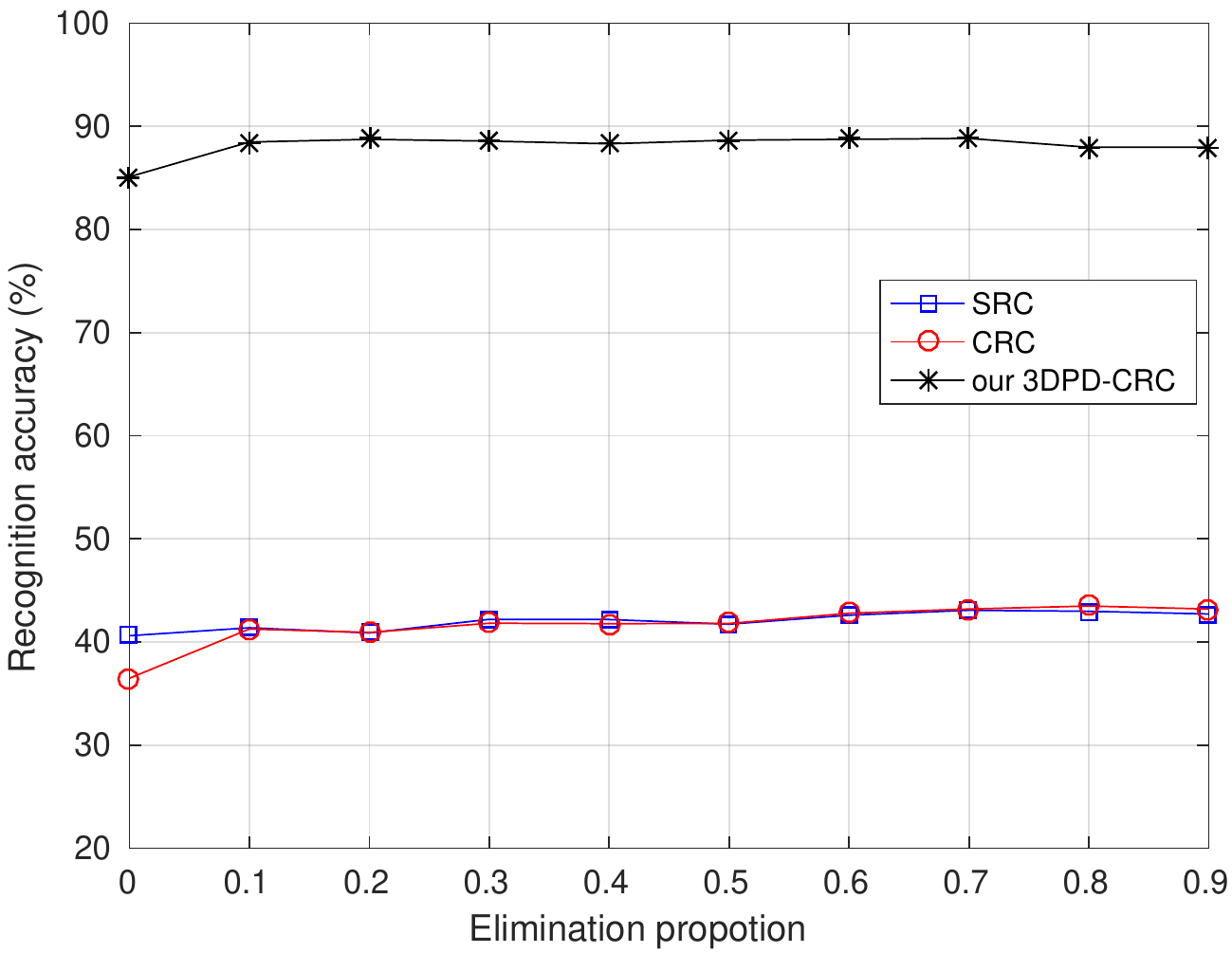}
	}
	\subfloat[4 training samples]{
		\label{fig7_c}
		\includegraphics[trim = 36mm 95mm 42mm 100mm, clip, width=0.32\linewidth]{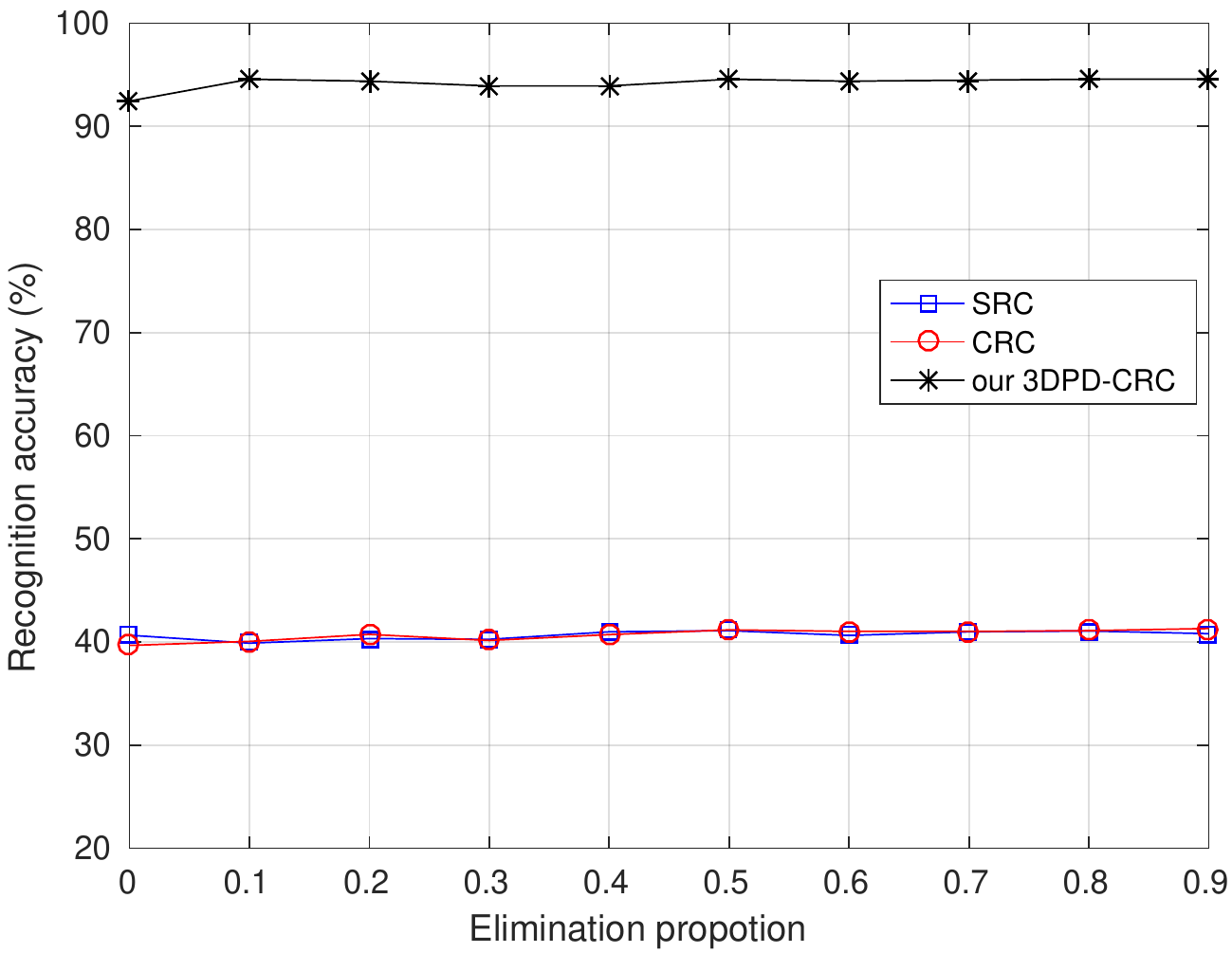}
	}
	\caption{Face recognition rates $(\%)$ of different methods with different randomly selected training samples on PIE}
	\label{fig7}
\end{figure*}
{\color{black}We used a similar split to construct training and test sets} for the PIE dataset.
The only difference here is that we rendered 4 virtual faces for each training face with $\pm 15^\circ$ and $\pm 30^\circ$ yaw rotations.

The results of SRC, CRC and our 3DPD-CRC are shown in Fig.~\ref{fig7}(a)-(c).
According to these figures, the proposed 3DPD-CRC method performs much better than SRC and CRC in terms of face classification {\color{black} accuracy} across all different sizes of training {\color{black} sets}.
It should be noted that the improvements {\color{black} achieved by} the proposed method on PIE and FERET datasets are much higher than that on the ORL dataset.
The main reason is that FERET and PIE contain more variations in appearance than ORL.
In such {\color{black} scenarios}, the superiority of our algorithm is more {\color{black} dramatic}.

\section{Conclusion}
In this paper, we proposed a dictionary integration algorithm using 3D morphable face models for pose-invariant CRC-based face classification.
The key innovation of the proposed method is to accomplish face recognition by utilizing 3DMM {\color{black} for} training data augmentation,
which makes CRC robust to pose variations.
The strength of the technique lies in successfully generating virtual faces with pose variations using 3DMM, 
and thereby enhancing the capacity {\color{black} of the dictionary to reconstruct input signals faithfully.}
Moreover, the extended dictionary is optimised {\color{black} on-line} using an elimination scheme, which further improves the accuracy of the proposed face classification algorithm.
We believe that our promising results will encourage more work on synthesising an informative dictionary and lead to successful solutions for other application domains in the future.

\bibliographystyle{IEEEtran}
\bibliography{IEEEabrv,egbib}
\end{document}